\pdfoutput=1

\documentclass[11pt]{article}

\usepackage{EMNLP2022}

\usepackage{times}
\usepackage{latexsym}

\usepackage[T1]{fontenc}

\usepackage[utf8]{inputenc}

\usepackage{microtype}

\usepackage{inconsolata}

\usepackage{subcaption}
\usepackage{graphicx}
\usepackage{booktabs}
\usepackage[normalem]{ulem}
\usepackage{multirow}
\usepackage{enumitem}

\usepackage{amsmath}
\DeclareMathOperator{\flop}{c}

\title{Efficient Large Scale Language Modeling with Mixtures of Experts}

\author{
{\bf Mikel Artetxe}\thanks{Equal contribution. Authors listed alphabetically.}{\normalfont ,}
{\bf Shruti Bhosale}\footnotemark[1]{\normalfont ,}
{\bf Naman Goyal}\footnotemark[1]{\normalfont ,}
{\bf Todor Mihaylov}\footnotemark[1]{\normalfont ,}
{\bf Myle Ott}\footnotemark[1]{\normalfont ,}
{\bf Sam Shleifer}\footnotemark[1]{\normalfont ,}\\
{\bf Xi Victoria Lin}, {\bf
Jingfei Du}, {\bf
Srinivasan Iyer}, {\bf
Ramakanth Pasunuru}, {\bf
Giri Anantharaman}, {\bf
Xian Li},\\ {\bf
Shuohui Chen}, {\bf
Halil Akin}, {\bf
Mandeep Baines}, {\bf
Louis Martin}, {\bf
Xing Zhou}, {\bf
Punit Singh Koura},\\ {\bf
Brian O'Horo}, {\bf
Jeff Wang}, {\bf
Luke Zettlemoyer}, {\bf
Mona Diab}, {\bf
Zornitsa Kozareva}, {\bf
Ves Stoyanov}\\
Meta AI
}
\begin{document}
\maketitle

\begin{abstract}
Mixture of Experts layers (MoEs) enable efficient scaling of language models through conditional computation. This paper presents a detailed empirical study of how autoregressive MoE language models scale in comparison with dense models in a wide range of settings: in- and out-of-domain language modeling, zero- and few-shot priming, and full-shot fine-tuning. With the exception of fine-tuning, we find MoEs to be substantially more compute efficient. At more modest training budgets, MoEs can match the performance of dense models using $\sim$4 times less compute. This gap narrows at scale, but our largest MoE model (1.1T parameters) consistently outperforms a compute-equivalent dense model (6.7B parameters). Overall, this performance gap varies greatly across tasks and domains, suggesting that MoE and dense models generalize differently in ways that are worthy of future study. We make our code and models publicly available for research use.\footnote{\url{https://github.com/pytorch/fairseq/tree/main/examples/moe_lm}} %
\end{abstract}

\section{Introduction}

Large Language Models (LMs) 
achieve remarkable accuracy and generalization ability when fine tuned for NLP tasks~\citep{peters2018deep,devlin2018bert,liu2019roberta,lan2020albert,raffel2020exploring}. They are also capable zero- and few-shot learners ~\citep{brown2020gpt3}, with the ability to generalize to tasks not seen during training.
A reliable way to improve LM accuracy in all of these settings is by scaling up: increasing the number of parameters and the amount of computation used during training and inference~\citep{raffel2020exploring,brown2020gpt3,fedus2021switch}. In fact, some generalization properties only emerge in very large models, including much improved zero- and few-shot learning~\citep{brown2020gpt3}.

Unfortunately, the corresponding growth in computational resources required to train state-of-the-art language models is a barrier for many in the research community~\cite{schwartz2019green}.
There is also a concern about the environmental costs associated with training and deploying such models~\citep{strubell2019energy,gupta2021chasing,bender2021dangers,patterson2021carbon} motivating research into more efficient model designs~\citep{lepikhin2021gshard,fedus2021switch,Lewis2021BASELS}.

\begin{figure}[t!]
\centering
\includegraphics[width=\linewidth]{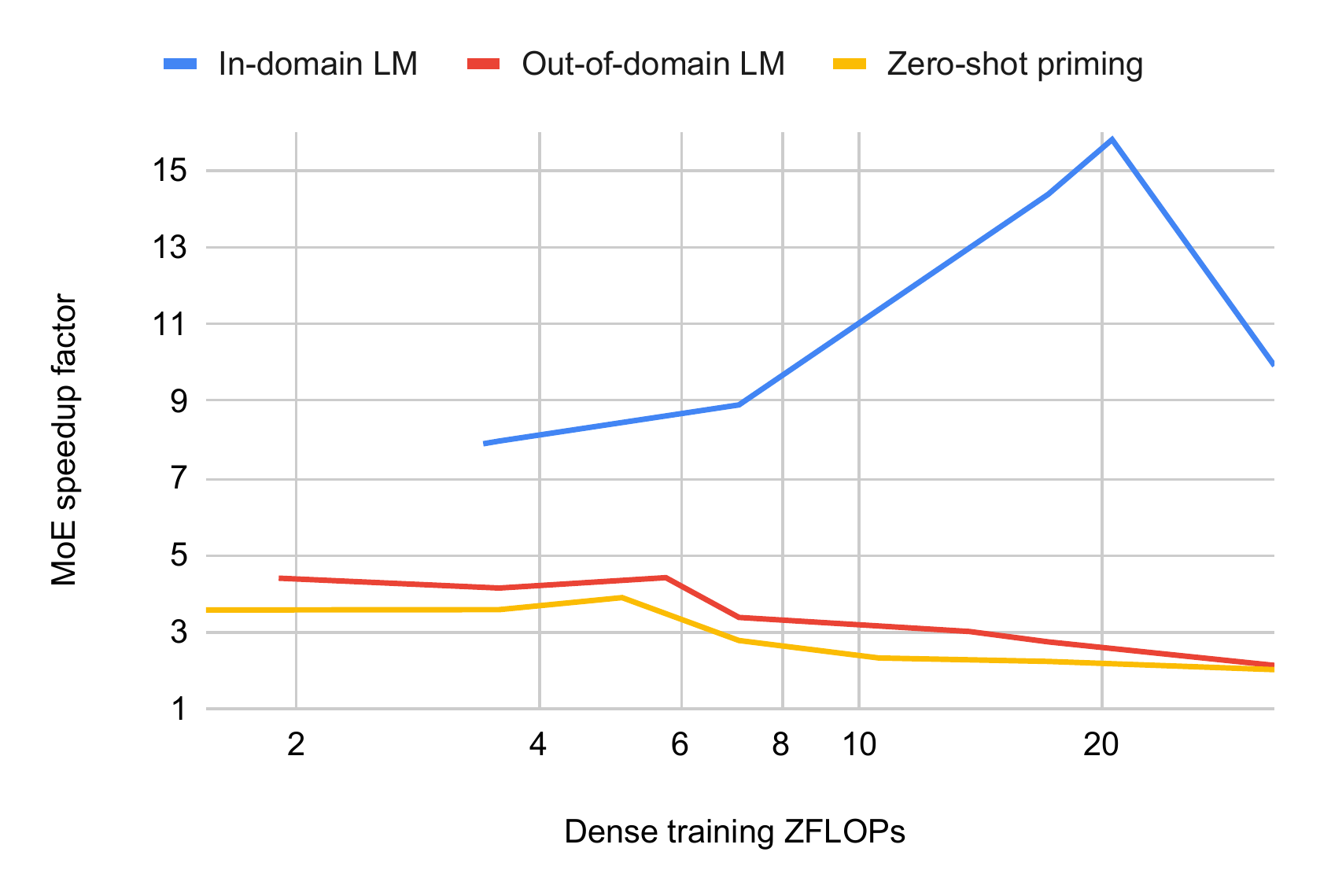}
\caption{
\textbf{Estimate of how much more efficient MoEs are relative to dense models.} A speedup factor of $y$ indicates that an MoE model can match the performance of the corresponding dense model---trained with $x$ ZFLOPs---using $y$ times less compute (i.e., $x/y$ ZFLOPs). We estimate this factor according to validation perplexity for in-domain language modeling, the Pile perplexity for out-of-domain language modeling, and average accuracy across 6 tasks for zero-shot priming. See \S\ref{subsubsec:efficiency-gain} for more details.
}
\label{fig:efficiency-gain}
\end{figure}

\emph{Sparse} models allow for increased number of learnable parameters without the associated computational costs. 
For example, sparsely gated mixture of experts (\emph{MoE})~\cite{lepikhin2021gshard} have been successfully used for language modeling and machine translation~\citep{lepikhin2021gshard,Lewis2021BASELS,roller2021hash}, but are yet to be shown effective for fine-tuning~\cite{fedus2021switch} as well as zero- and few-shot learning. 
We hypothesize that sparse models are comparatively accurate to dense models but at a much lower computational footprint. 
To measure this claim, we train traditional dense and MoE language models ranging in size from several hundred million parameters to more than one trillion parameters and present a careful empirical comparison of these models on downstream tasks in zero-shot, few-shot and fully supervised settings.

As shown in Figure \ref{fig:efficiency-gain}, we find that MoE models can indeed achieve similar downstream task performance as dense models at a fraction of the compute. For models with relatively modest compute budgets, a MoE model can perform on par with a dense model that requires almost four times as much compute.
Downstream task performance improves with scale for both MoE models and dense models.
While we observe that the performance gap narrows as we increase model size, even at larger compute budgets ($\sim$ 5000 GPU days), our largest MoE model (1.1T parameters) outperforms a dense model with similar computational cost (6.7B parameters).
We further compare and contrast the performance of dense and sparse models with similar computational signatures and observe some performance variations across tasks and domains, suggesting this an interesting area for future research. In summary, our contributions are:
\begin{itemize}[leftmargin=*]
    \item We present a comprehensive study of sparse models for zero and few-shot learning at scale; 
    \item We demonstrate that even at scale sparse MoE models can yield competitive zero and few-shot performance at a fraction of the computation for model training and inference;
    \item We observe some differences in how dense and sparse models generalize at scale suggesting complementary behaviour that could be an interesting future research direction.
\end{itemize}

\section{Background and Related Work}

\subsection{Large Language Models / GPT-3}

Progress in the field of NLP has been driven by increasingly large Language Models (LMs) pretrained on large text datasets. While numerous variations have been proposed, such LMs are predominantly based on the transformer architecture~\citep{vaswani2017attention}. Models are pretrained by hiding parts of the input: predicting the next word sequentially left-to-right, masking words in the text~\citep{devlin2018bert,liu2019roberta}, or perturbing and/or masking spans~\citep{lewis-etal-2020-bart,raffel2020exploring}. The resulting models can be quickly adapted to perform new tasks at high accuracy by fine-tuning on supervised data~\citep{devlin2018bert,liu2019roberta}.

Recently, GPT-3~\citep{brown2020gpt3} demonstrated that large LMs can perform zero- and few-shot learning without fine-tuning through in-context learning. Notably, many of these in-context zero- and few-shot learning behaviors emerge or amplify at scale. Concurrent to our work, \citet{rae2022scaling} and \citet{smith2022using} further explore scaling dense language models. %

\subsection{Sparse models}
One drawback of dense model scaling is that it grows increasingly computationally expensive. To more efficiently increase model capacity, conditional compute strategies have been developed \citep{bengio2013estimating,davis2013low,cho2014exponentially,bengio2015conditional}, where each input  activates a subset of the model. Recent work \citep{Lewis2021BASELS,lepikhin2021gshard,fedus2021switch,fan2021beyond} has studied different conditional compute strategies that work well with Transformer models for natural language tasks. In this work, we focus on Sparsely Gated Mixture of Expert (MoE) models \citep{shazeer2017outrageously,lepikhin2021gshard}. Sparse MoE models replace the dense feed forward network block in every alternate Transformer layer with an MoE layer. The MoE layer has a routing gate that learns which tokens are to be mapped to which set of experts (we use top-2 experts). To ensure scalability and training efficiency, it is also common to include a weighted gate loss term as in \citet{lepikhin2021gshard} to the cross entropy loss to encourage the tokens to be uniformly distributed to the experts. 
Concurrent to our work, \citet{du2021glam}, \citet{rajbhandari2022deepspeedmoe} and \citet{clark2022unified} also study MoE scaling.

\begin{table*}[ht]
\begin{center}
\begin{small}
\addtolength{\tabcolsep}{-2.5pt}
\begin{tabular}{crrrrrrrrrrrrrrrrrr}
\toprule
& \multicolumn{5}{c}{GPT-3 (dense)} && \multicolumn{5}{c}{Ours (dense)} && \multicolumn{5}{c}{Ours (MoE)} & \\
\cmidrule{2-6} \cmidrule{8-12} \cmidrule{14-18}
& \multicolumn{1}{c}{\emph{size}} & \multicolumn{1}{c}{\emph{cost}} & \multicolumn{1}{c}{$l$} & \multicolumn{1}{c}{$h$} & \multicolumn{1}{c}{$e$} &
& \multicolumn{1}{c}{\emph{size}} & \multicolumn{1}{c}{\emph{cost}} & \multicolumn{1}{c}{$l$} & \multicolumn{1}{c}{$h$} & \multicolumn{1}{c}{$e$} &
& \multicolumn{1}{c}{\emph{size}} & \multicolumn{1}{c}{\emph{cost}} & \multicolumn{1}{c}{$l$} & \multicolumn{1}{c}{$h$} & \multicolumn{1}{c}{$e$} &
\\
\midrule
& 125M & 0.36 & 12 & 768 & -- &
& 125M & 0.36 & 12 & 768 & -- &
& 15B & 0.43 & 12 & 768 & 512 &
\\
& 355M & 1.06 & 24 & 1024 & -- &
& 355M & 1.06 & 24 & 1024 & -- &
& 52B & 1.30 & 24 & 1024 & 512 &
\\
& 760M & 2.13 & 24 & 1536 & -- &
& \multicolumn{5}{c}{---} &
& \multicolumn{5}{c}{---} &
\\
& 1.3B & 3.57 & 24 & 2048 & -- &
& 1.3B & 3.57 & 24 & 2048 & -- &
& 207B & 4.53 & 24 & 2048 & 512 &
\\
& 2.7B & 7.08 & 32 & 2560 & -- &
& 2.7B & 7.08 & 32 & 2560 & -- &
& \multicolumn{5}{c}{---} &
\\
& 6.7B & 17.12 & 32 & 4096 & -- &
& 6.7B & 17.12 & 32 & 4096 & -- &
& 1.1T & 22.27 & 32 & 4096 & 512 &
\\
& 13B & 32.67 & 40 & 5120 & -- &
& 13B & 32.67 & 40 & 5120 & -- &
& \multicolumn{5}{c}{---} &
\\
& 175B & 430.17 & 96 & 12288 & -- &
& \multicolumn{5}{c}{---} &
& \multicolumn{5}{c}{---} &
\\
\bottomrule
\end{tabular}%
\end{small}
\end{center}
\caption{\textbf{Dense and mixture of expert (MoE) model details}. \emph{size}: number of parameters, \emph{cost}: training ZFLOPs, $l$: layers, $h$: hidden dimension, $e$: number of experts. All models are trained for 300B tokens with a sequence length of 2048 tokens. Models within the same row are roughly comparable. We estimate the training cost in ZFLOPs analytically (see Appendix~\ref{app:flops}).}
\label{tab:models}
\end{table*}

\subsection{Zero-shot and Few-shot Learning} \label{subsec:background_fewshot}

Recent works \citep{schick-schutze-2021-exploiting,radford2019language} have directly evaluated LMs on unseen tasks successfully (zero-shot learning), by recasting their task inputs as cloze-style prompt completion tasks. This is in contrast to the traditional approach of augmenting LMs with task-specific heads, followed by supervised fine-tuning \citep{devlin2018bert,raffel2020exploring}. Subsequently, \citet{brown2020gpt3} demonstrated that priming LMs with a few input-output examples (few-shot learning) before careful prompting can improve task performance, that grows with model scale without any fine-tuning, and this gave rise to new resources for prompt engineering \citep{bach2022promptsource}. In this paper, we contrast the zero-shot, few-shot, and fully supervised fine-tuning performance of dense and MoE models. Finally, \citet{schick2021smalllanguagemodels} perform few-shot learning by few-shot fine-tuning using pattern-exploiting training, whose efficiency can be improved by performing partial fine-tuning of a small number of additional task-specific parameters instead \citep{lester2021power,li2021prefix,houlsby2019parameter}. %

\subsection{Large-scale training}

Many of the models we consider in this work are too big to be trained using standard data parallel techniques, since parameter storage would exceed the usable memory of a single GPU.
We adopt several techniques to make these models feasible to train, including pure FP16 training, activation checkpointing and fully sharded data parallel training. These techniques are described in more depth in Appendix~\ref{app:scaling}.

\section{Experimental Setup}

\subsection{Models}

We train autoregressive (decoder-only) transformer models that roughly match the sizes and architecture explored in~\citet{brown2020gpt3}. Model sizes are summarized in Table~\ref{tab:models}.
We use pre-normalization transformer blocks~\citep{baevski2018adaptive,child2019generating} and GELU activations~\citep{hendrycks2016gelu}.
We differ from \citet{brown2020gpt3} in two ways: (1) we use only dense attention, while they alternate between dense and locally banded sparse attention; and (2) we train our models with sinusoidal positional embeddings, following Shortformer~\cite{press2020shortformer}.\footnote{Early experiments found this to produce comparable results with fewer learned parameters.}

We also train MoE models that mirror our dense model configurations (see the third set of columns in Table~\ref{tab:models}), so that comparisons are approximately matched in terms of the number of floating point operations (\emph{FLOP}s).
Our MoE models follow the design proposed in \citet{lepikhin2021gshard} with alternating dense and expert layers and top-2 expert selection.
We use 512 experts in each expert layer ($E=512$).
Each expert has a \emph{capacity} of $\frac{C \cdot B}{E}$ tokens, where $C$ is a \emph{capacity factor} that we set to $2$ and $B$ is the total batch size in tokens. Capacity refers to the maximum number of tokens that are routed to each expert.
Once an expert is at capacity for a given batch, additional tokens are considered to be ``overflowed" with their representations passed-through via the residual connection.

\citet{fedus2021switch} report instability training large MoE models and suggest rescaling the initial model weights, which we do not find necessary.
We instead observe that expert parameters have an $E$-times smaller batch size relative to dense (data parallel) parameters and accordingly rescale expert gradients by a factor $\frac{1}{\sqrt{E}}$.
This rescaling aligns with theory suggesting that an $E$-times increase in batch size should be accompanied by a $\sqrt{E}$ increase in learning rate~\citep{krizhevsky2014one}.

Following \citet{brown2020gpt3}, we train our models for 300B tokens\footnote{While we control the total number of tokens to be the same as \citet{brown2020gpt3}, our pretraining data is not the same. See \S\ref{subsec:pretraining_data} for further details.}
with a context size (sequence length) of 2048 tokens.
The batch size and learning rate are set according to the model size following \citet{brown2020gpt3}.
We linearly warm-up the learning rate from $0$ over the first 375M tokens and linearly decay back to $0$ over the remaining tokens.
We use the Adam optimizer~\cite{kingma2014adam} with $\beta_1=0.9$, $\beta_2=0.98$, $\epsilon=10^{-8}$, weight decay of 0.01 and dropout of 0.1.\footnote{We note that our 355M dense and 52B MoE models (FLOPs-matched) were trained without dropout, which we find slightly improves performance at smaller scale.}

We train our models in PyTorch~\cite{paszke2017automatic} using \textsc{fairseq}~\citep{ott2019fairseq}.

\subsection{Pretraining data}
\label{subsec:pretraining_data}

We pretrain our models on a union of six English-language datasets, including the five datasets used to pretrain RoBERTa~\citep{liu2019roberta} and the English subset of CC100, totalling 112B tokens corresponding to 453GB:
\begin{itemize}[leftmargin=*]
\item \textbf{BookCorpus}~\citep{zhu2015bookcorpus} consists of more than 10K unpublished books (4GB);
\item \textbf{English Wikipedia}, excluding lists, tables and headers (12GB);
\item \textbf{CC-News}~\citep{nagel2016ccnews} contains 63 millions English news articles crawled between September 2016 and February 2019 (76GB);
\item \textbf{OpenWebText}~\citep{gokaslan2019openwebtext}, an open source recreation of the WebText dataset used to train GPT-2 (38GB);
\item \textbf{CC-Stories}~\citep{trinh2018simple} contains a subset of CommonCrawl data filtered to match the story-like style of Winograd schemas (31GB);
\item \textbf{English CC100}~\citep{wenzek-etal-2020-ccnet}, a dataset extracted from CommonCrawl snapshots between January 2018 and December 2018, filtered to match the style of Wikipedia (292GB).
\end{itemize}
We encode our data using the same Byte-Pair Encoding (BPE) as GPT-2~\citep{radford2019language} and RoBERTa~\citep{liu2019roberta} with a vocabulary of 50K subword units.

\subsection{Evaluation}

We evaluate models in terms of their in-domain and out-of-domain perplexity, as well as downstream task performance. 

\subsubsection{Perplexity Evaluation}

We first evaluate our models on their ability to predict the next token in a sequence as measured by perplexity. Similar to training, we concatenate all documents in a given dataset using empty lines as separators, split the resulting sequence into non-overlapping blocks of 2048 tokens, and score each block independently.\footnote{One limitation of this approach is that the first tokens in each block have limited context, as they do not condition on tokens from preceding blocks. Although more expensive, better results could be obtained using a sliding window approach. Nevertheless, this form of chunking the input is standard in language model evaluation.}

We evaluate and report perplexity in both \textbf{in-domain} and \textbf{out-of-domain} settings.
In-domain, we sample a held-out subset of the combined pretraining data (\S\ref{subsec:pretraining_data}).
For out-of-domain we use data from The Pile \citep{gao2021thepile}, a public dataset that combines data from 22 diverse sources (e.g., ArXiv, Github, OpenSubtitles, etc.).
We report perplexities on the official test set of each individual subset, as well as the average across all subsets.

\subsubsection{Downstream Evaluation} \label{subsec:downstream}

We target models that can perform downstream tasks well. Recent work shows that good perplexity performance does not always align with good performance on downstream tasks~\cite{tay2021scale}. Hence, we evaluate our models accordingly.

\paragraph{Benchmarks.}
We evaluate our models on a subset of the tasks considered in \citet{brown2020gpt3}.
As GPT-3 performance varies greatly across tasks and model sizes, we focus on tasks for which GPT-3 either demonstrated consistent gains from scaling, or consistent gains going from zero-shot to few-shot settings.

\textbf{Few-shot:} we use WinoGrande~\citep{sakaguchi2020winogrande}, StoryCloze~\citep{mostafazadeh-etal-2016-corpus} and OpenBookQA~\citep{mihaylov-etal-2018-suit}, the only non-generation tasks for which \citet{brown2020gpt3} reported meaningful gains over zero-shot at our scale.\footnote{Defined as an improvement of at least 2 accuracy points over zero-shot learning and the majority class baseline for at least one GPT-3 model no bigger than 6.7B.}  We exclude SuperGLUE, since we were not able to reproduce results reported in \citet{brown2020gpt3} using the public GPT-3 API.\footnote{Different from other tasks, we were not able to reproduce GPT-3 results on SuperGLUE using the OpenAI API and our evaluation protocol. The authors confirmed that they used a different evaluation protocol for SuperGLUE through personal correspondence.}

\textbf{Zero-shot:} in addition to the 3 few-shot tasks, we evaluate on ReCoRD~\citep{zhang2018record}, HellaSwag~\citep{zellers-etal-2019-hellaswag} and PIQA~\citep{bisk2020piqa}. \citet{brown2020gpt3} reported strong results and monotonic improvements from scaling on these tasks.

\paragraph{Evaluation protocol.} Following \citet{brown2020gpt3}, we report results on the development set for all tasks except OpenBookQA and StoryCloze, for which we use the test set. For few-shot learning, we report the average results across 25 runs, randomly sampling a different set of few-shot examples from the training set each time.\footnote{StoryCloze does not have a training set, so we follow \citet{brown2020gpt3} and sample few-shot examples from the development set instead.} For priming, we further shuffle the few-shot examples for each test instance. Following \citet{brown2020gpt3}, we use k=50 few-shot examples for WinoGrande, k=70 for StoryCloze and k=100 for OpenBookQA.
In cases where this exceeds the maximum context length for the model, we truncate the prompt keeping the maximum number of full examples that fit.

\paragraph{Baselines.} We compare to the published GPT-3 numbers \citep{brown2020gpt3} as our primary baseline. To validate our experimental framework, we also evaluate GPT-3 leveraging the OpenAI API using our own evaluation code and settings. Unfortunately, the correspondence between model sizes and model names in the OpenAI API is not published. We follow other published work \citep{gao2021thepile} and guess the correspondence based on our results from the public API as compared to results in \citet{brown2020gpt3} (see \S\ref{subsec:zero_shot_results}).

\paragraph{Methods.} We compare both priming and fine-tuning-based approaches.
\begin{itemize}[leftmargin=*]
\item \textbf{Priming:} We use a language model to separately score each label choice using the same templates as \citet{brown2020gpt3}, and pick the one with the highest score. For few-shot learning, we use a single newline to separate examples. Our scoring function follows the description in \citet{brown2020gpt3}:
\begin{itemize}
\item{\bf For WinoGrande}, we take the log-likelihood of the common suffix of the different candidates.
\item{\bf For OpenBookQA}, we normalize by the unconditional probability of each candidate by taking $\frac{p(\mathtt{completion}|\mathtt{context})}{p(\mathtt{completion}|\mathtt{answer\_context})}$, where we use the string \textit{``Answer: ''} as answer\_context.
\item{\bf For ReCoRD}, we take the sum of per-token log-probabilities.\footnote{This is different from \citet{brown2020gpt3}, who take the average per-token log-probability for this task. This worked worse in our preliminary experiments.}
\item{\bf For all the other tasks}, we take the average of per-token log-probabilities, ignoring the common prefix of the different candidates.
\end{itemize}
\item \textbf{Fine-tuning:} Although supervised fine-tuning of pre-trained LMs on task specific training data, $\mathcal{D}$, requires updating and storage of all model parameters per task, the process typically produces significant task specific performance improvements. We contrast the fine-tuning performance of sparse models and their dense counterparts following \cite{radford2018gpt}, which applies an additional task-specific linear layer $W_y$ on the representation from the final transformer block for each input candidate separately, followed by a softmax layer.
We fine-tune all model parameters using the entire training set (fully supervised learning). In addition to our zero-shot tasks, we also evaluate on 3 widely-used classification tasks: BoolQ~\citep{clark-etal-2019-boolq}, MNLI~\citep{williams-etal-2018-broad} and SST-2~\citep{socher-etal-2013-recursive}. More details are in Appendix \ref{sec:fine_tuning_settings}.

\end{itemize}

\begin{figure*}[t]
     \centering
     \begin{subfigure}[b]{0.485\textwidth}
         \centering
         \includegraphics[width=\textwidth]{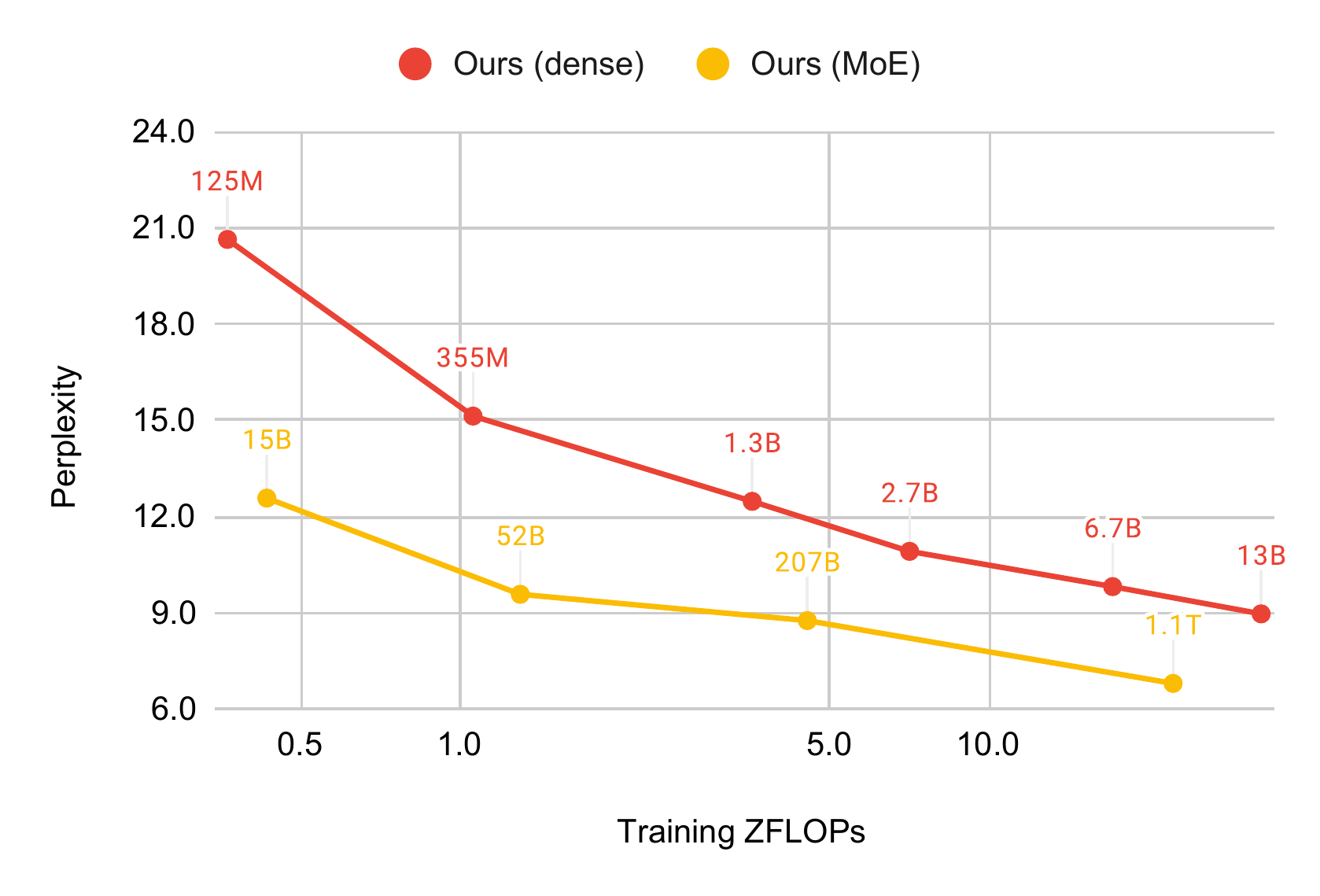}
         \caption{In-domain (validation)}
         \label{fig:ppl-valid}
     \end{subfigure}
     \hfill
     \begin{subfigure}[b]{0.485\textwidth}
         \centering
         \includegraphics[width=\textwidth]{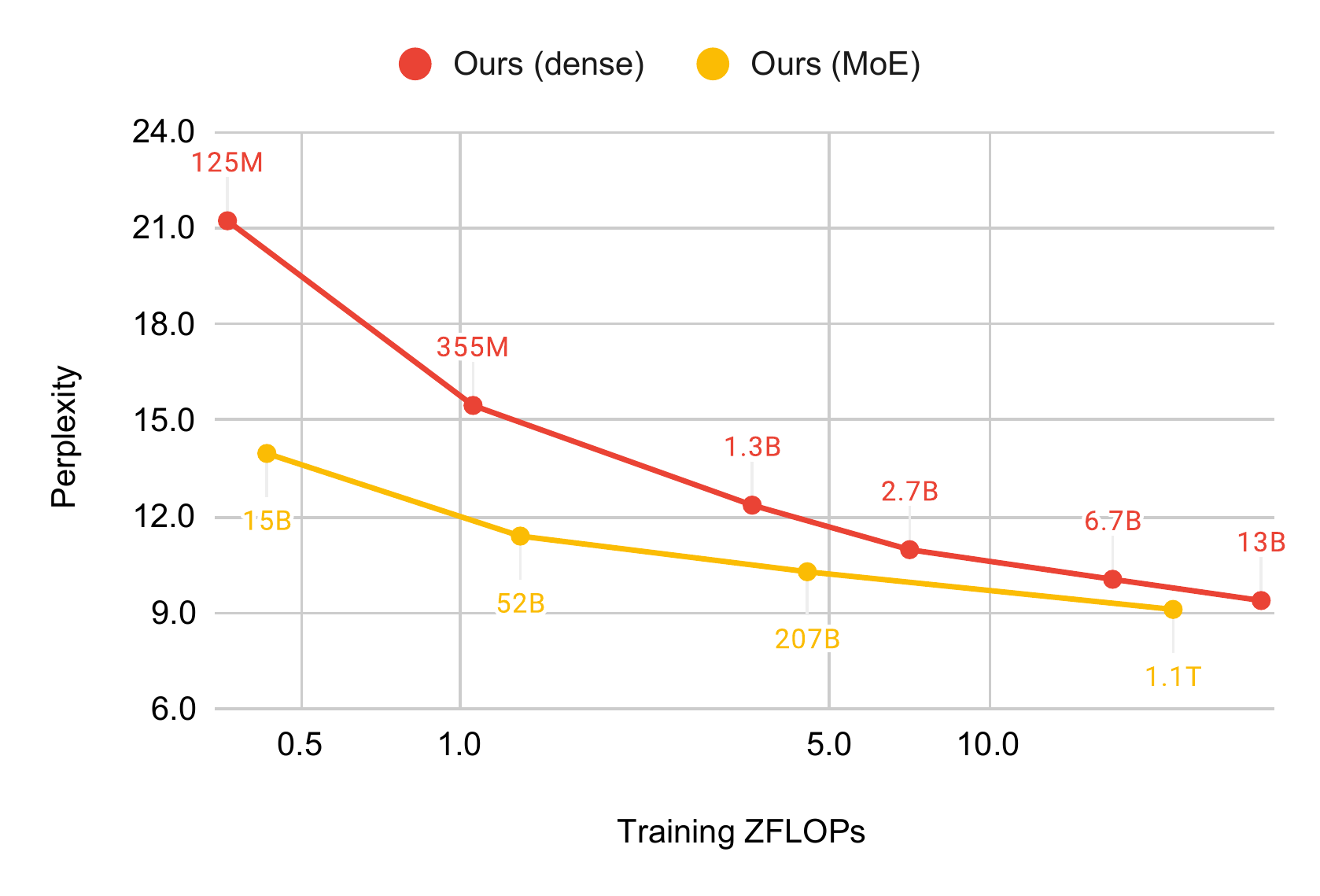}
         \caption{Out-of-domain (the Pile)}
         \label{fig:ppl-thepile}
     \end{subfigure}
     \hfill
    \caption{\textbf{Language modeling perplexity.} For the Pile, we report the average perplexity across the 22 subsets.}
    \label{fig:ppl}
\end{figure*}

\begin{figure*}[t]
     \centering
     \begin{subfigure}[b]{0.485\textwidth}
         \centering
         \includegraphics[width=\textwidth]{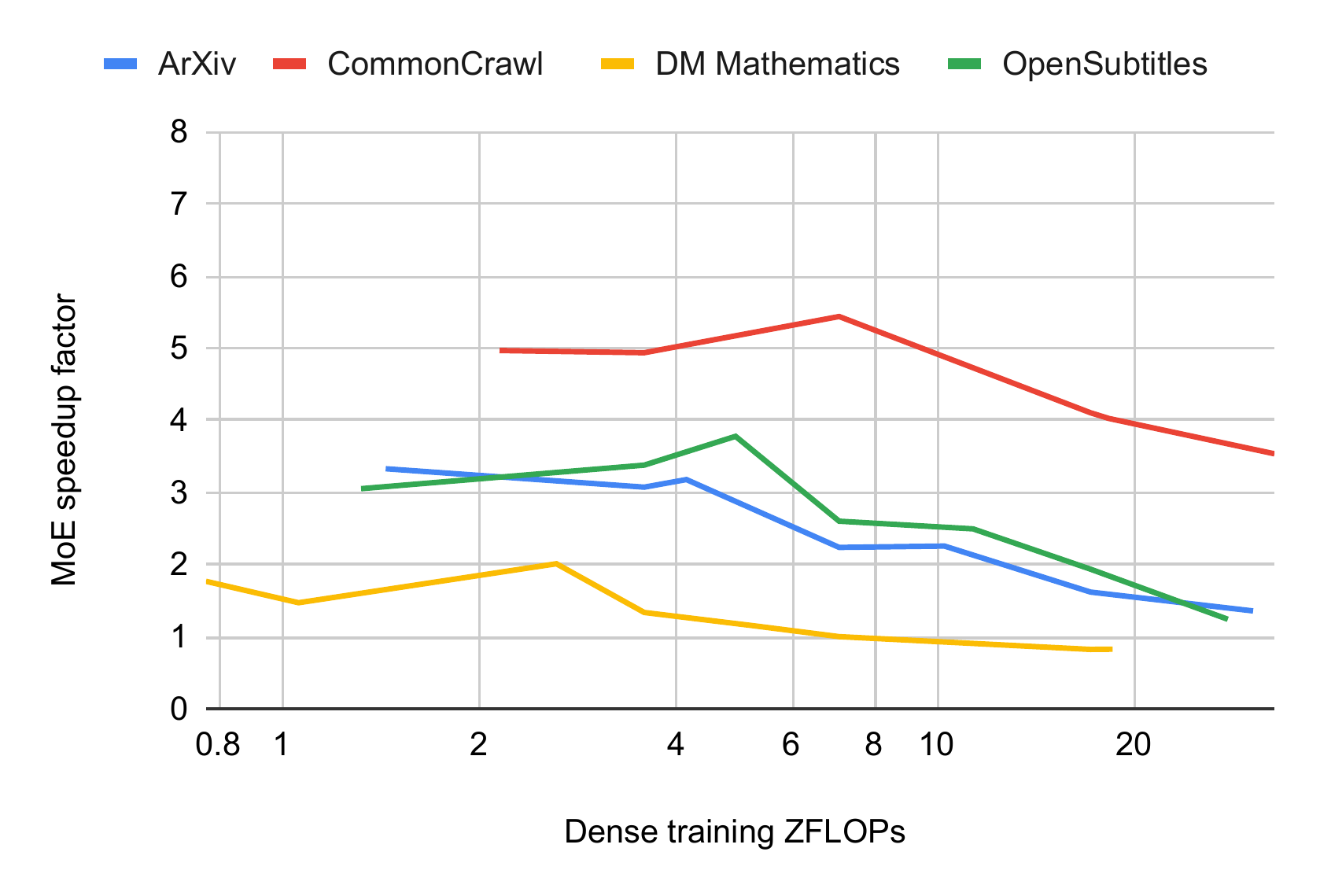}
         \caption{Language modeling (the Pile)}
         \label{fig:efficiency-gain-ppl}
     \end{subfigure}
     \hfill
     \begin{subfigure}[b]{0.485\textwidth}
         \centering
         \includegraphics[width=\textwidth]{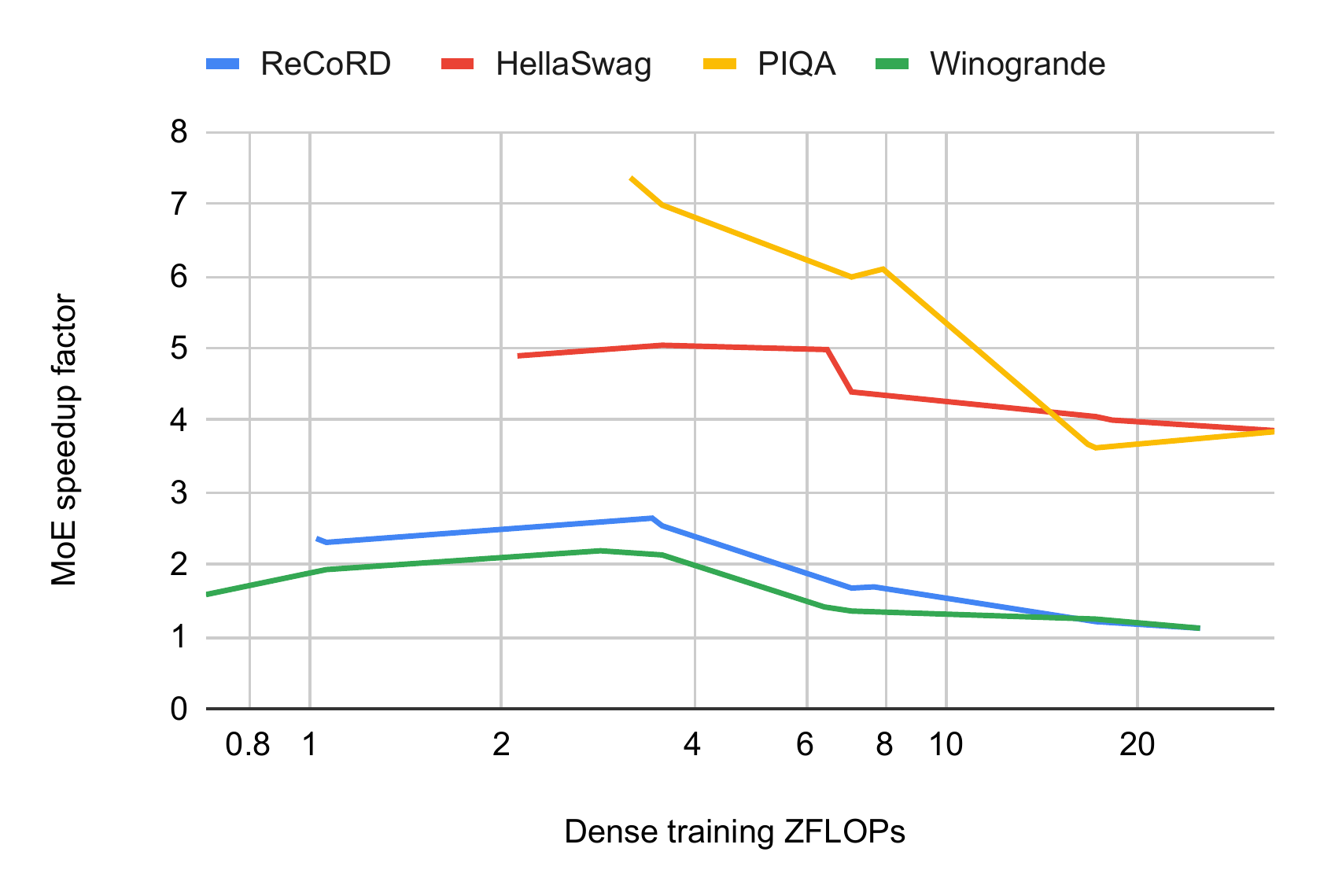}
         \caption{Zero-shot priming}
         \label{fig:efficiency-gain-zeroshot}
     \end{subfigure}
     \hfill
    \caption{
    \textbf{Estimate of how much more efficient MoEs are relative to dense models in representative datasets.} A speedup factor of $y$ indicates that an MoE model can match the performance of the corresponding dense model using $y$ times less compute. Refer to \S\ref{subsubsec:efficiency-gain} for more details.
    }
    \label{fig:efficiency-gain-datasets}
\end{figure*}

\subsubsection{MoE speedup factor}
\label{subsubsec:efficiency-gain}

We hypothesize that sparse models can achieve comparable performance at a smaller compute budget. As such, it is informative to measure how much more efficient MoEs are at achieving a specific performance level relative to dense models. We estimate how many FLOPs $\flop (t)$ the model needs to achieve performance $t$ in a particular task (as measured by perplexity for language modeling and accuracy for downstream tasks) using either an MoE or a dense model. Given that we only have  discrete observations, we estimate exact missing values by interpolating on a logarithmic scale as follows:
$$ \flop(t) = \exp \left( \log \flop_{lo}(t) + r \left(\log \flop_{hi}(t) - \log \flop_{lo}(t) \right) \right)$$
where $r = \frac{t - t_{lo}}{t_{hi} - t_{lo}}$, $t_{lo}$ and $t_{hi}$ are the closest performance to $t$ from the available models while being lower and higher than $t$, respectively, and $\flop_{lo}(t)$ and $\flop_{hi}$ are their corresponding training cost in ZFLOPs.

The interpolation gives us matching performance levels for dense and MoE models. We use them to compute the MoE speedup factor $\flop_{dense}(t) / \flop_{moe}(t)$. For example, if a dense model requiring 20 ZFLOPs achieves a performance of $90\%$ on a given task and a MoE model requiring 5 ZFLOPs achieves the same performance, then the formula produces saving factor of 4. We visualize the savings curve using $\flop_{dense}(t)$ in the $x$ axis, which allows us to contrast speedup in different tasks in a comparable scale.

\begin{table}[t]
\begin{center}
\begin{small}
\addtolength{\tabcolsep}{-2.5pt}
\resizebox{0.48\textwidth}{!}{
\begin{tabular}{cr|cccccc|c}
\toprule
&& RE & HS & PI & WG & SC & OB & avg \\
\midrule
\multirow{8}{*}{\shortstack{GPT-3 \\ (paper)}}
& 125M & 70.8 & 33.7 & 64.6 & 52.0 & 63.3 & 35.6 & 53.3 \\
& 355M & 78.5 & 43.6 & 70.2 & 52.1 & 68.5 & 43.2 & 59.4 \\
& 760M & 82.1 & 51.0 & 72.9 & 57.4 & 72.4 & 45.2 & 63.5 \\
& 1.3B & 84.1 & 54.7 & 75.1 & 58.7 & 73.4 & 46.8 & 65.5 \\
& 2.7B & 86.2 & 62.8 & 75.6 & 62.3 & 77.2 & 53.0 & 69.5 \\
& 6.7B & 88.6 & 67.4 & 78.0 & 64.5 & 77.7 & 50.4 & 71.1 \\
& 13B & 89.0 & 70.9 & 78.5 & 67.9 & 79.5 & 55.6 & 73.6 \\
& 175B & 90.2 & 78.9 & 81.0 & 70.2 & 83.2 & 57.6 & 76.9 \\
\midrule
\multirow{4}{*}{\shortstack{GPT-3 \\ (API)}}
& ada & 77.4 & 42.9 & 70.3 & 52.9 & 68.6 & 41.0 & 58.9 \\
& babb. & 83.1 & 55.1 & 74.5 & 59.4 & 73.3 & 45.6 & 65.2 \\
& curie & 87.1 & 67.8 & 77.1 & 64.3 & 77.7 & 50.8 & 70.8 \\
& davi. & -- & 78.8 & 80.0 & 70.0 & 83.1 & 58.8 & -- \\
\midrule
\multirow{6}{*}{\shortstack{Ours \\ (dense)}}
& 125M & 69.3 & 33.7 & 65.3 & 52.1 & 66.0 & 35.4 & 53.6 \\
& 355M & 78.1 & 46.2 & 70.6 & 54.2 & 71.0 & 42.0 & 60.4 \\
& 1.3B & 83.5 & 58.4 & 74.6 & 58.1 & 76.8 & 49.4 & 66.8 \\
& 2.7B & 85.8 & 65.9 & 76.6 & 61.4 & 78.2 & 49.6 & 69.6 \\
& 6.7B & 87.5 & 70.2 & 78.2 & 64.7 & 80.5 & 51.8 & 72.2 \\
& 13B & 88.5 & 73.7 & 79.0 & 67.6 & 80.9 & 55.4 & 74.2 \\
\midrule
\multirow{4}{*}{\shortstack{Ours \\ (MoE)}}
& 15B & 77.8 & 53.2 & 74.3 & 53.4 & 73.6 & 42.0 & 62.4 \\
& 52B & 83.4 & 64.9 & 76.8 & 57.4 & 75.9 & 51.0 & 68.2 \\
& 207B & 86.0 & 70.5 & 78.2 & 60.9 & 78.1 & 50.8 & 70.7 \\
& 1.1T & 88.0 & 78.6 & 80.3 & 66.4 & 81.8 & 55.2 & 75.0 \\
\bottomrule
\end{tabular}
}
\end{small}
\end{center}
\caption{
\textbf{Zero-shot priming accuracy.}
\textit{GPT-3 (paper)} results taken from \citet{brown2020gpt3}, all the other results were obtained by us as described in \S\ref{subsec:downstream}.
\texttt{RE}: ReCoRD, \texttt{HS}: HellaSwag, \texttt{PI}: PIQA, \texttt{WG}: WinoGrande, \texttt{SC}: StoryCloze, \texttt{OB}: OpenBookQA.
We do not evaluate the largest GPT-3 model (davinci) on \texttt{RE} given the high price.
}
\label{tab:zeroshot}
\end{table}

\begin{figure}[t]
\centering
\includegraphics[width=\linewidth]{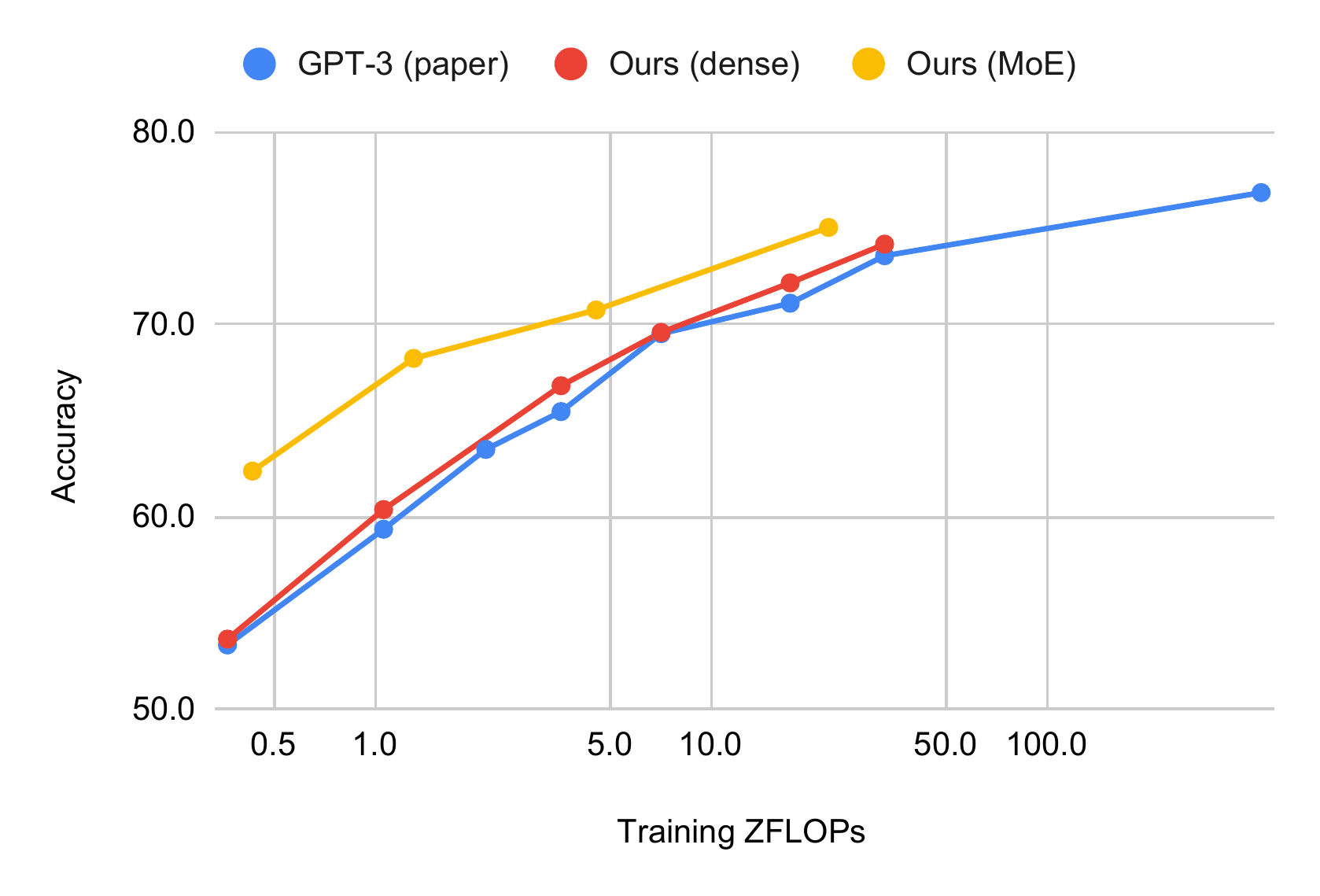}
\caption{
\textbf{Zero-shot priming accuracy averaged across 6 tasks as a function of compute cost.} Each point corresponds to a different, fully-trained model (see Table \ref{tab:models}).
\textit{GPT-3 (paper)} results taken from \citet{brown2020gpt3}.
}
\label{fig:zeroshot}
\end{figure}

\section{Results and Analysis}\label{sec:results}

\subsection{Language modeling perplexity}

We report our perplexity results in Figure \ref{fig:ppl}, and visualize the speedup curves in representative subsets of the Pile \citep{gao2021thepile} in Figure \ref{fig:efficiency-gain-ppl}. Refer to Appendix \ref{app:full-results} for full results for all the 22 subsets of the Pile.

We observe that all MoE models outperform their dense counterparts in all datasets, but their advantage greatly varies across domains and models. MoEs are most efficient when evaluated in-domain, where they are able to match the performance of dense models trained with 8-16x more compute (see Figure \ref{fig:efficiency-gain}). The improvement is more modest in out-of-domain settings, bringing a speedup of 2-4 on the Pile. This is reflected in Figure \ref{fig:ppl}, where the gap between the MoE and dense curves is substantially smaller in out-of-domain settings. Moreover, the advantage of MoEs over dense models decreases at scale: MoEs need $\sim$4 times less compute to match the performance of dense models trained with 2-6 ZFLOPs, but the speedup is $\sim$2 for dense models trained with $\sim$30 ZFLOPs. %

We also observe large difference across the subsets of the Pile, which correspond to different domains. As shown in Figure \ref{fig:efficiency-gain-ppl}, MoEs obtain the largest speedups in subsets that are closest to the training corpus (e.g., CommonCrawl). The efficiency gains are more moderate but still remarkable for other domains like ArXiv and OpenSubtitles. Our largest MoE model barely outperforms its dense counterpart on DM Mathematics (7.63 vs. 7.66 perplexity), which is arguably very different from the training domain.

\subsection{Downstream task evaluation}

\subsubsection{Zero-shot learning}\label{subsec:zero_shot_results}

We report zero-shot results in Table \ref{tab:zeroshot}, and visualize how the different model families scale in Figure \ref{fig:zeroshot}.

Our dense models perform at par with their GPT-3 counterparts. This is consistent across different tasks, with our models doing marginally better on average. We are thus able to match \citet{brown2020gpt3} despite some notable differences in our setup (e.g., different training corpus), establishing a solid baseline to evaluate MoE models on downstream tasks. Similarly, when using our own code to evaluate the strongest GPT-3 API backend (\textit{davinci}), we obtain numbers that replicate those reported in the original paper for their largest model, which reinforces that our evaluation settings are comparable to \citet{brown2020gpt3}.\footnote{We assume that \textit{ada} corresponds to the 355M model, \textit{babbage} corresponds to the 1.3B model, and \textit{curie} corresponds to the 6.7B model based on the API evaluation results.}

As with language modeling, MoEs outperform their dense counterparts for all datasets and model sizes. But, once again, we find the advantage narrows at scale as illustrated in Figure \ref{fig:zeroshot}. Similar to the domain differences in language modeling, we observe differences across downstream tasks. As shown in Figure \ref{fig:efficiency-gain-zeroshot}, MoEs obtain significant speedups in certain tasks like HellaSwag and PIQA, but this improvement is more modest in other tasks such as ReCoRD and Winogrande.

\subsubsection{Few-shot learning}

\begin{figure}[t!]
\centering
\includegraphics[width=\linewidth]{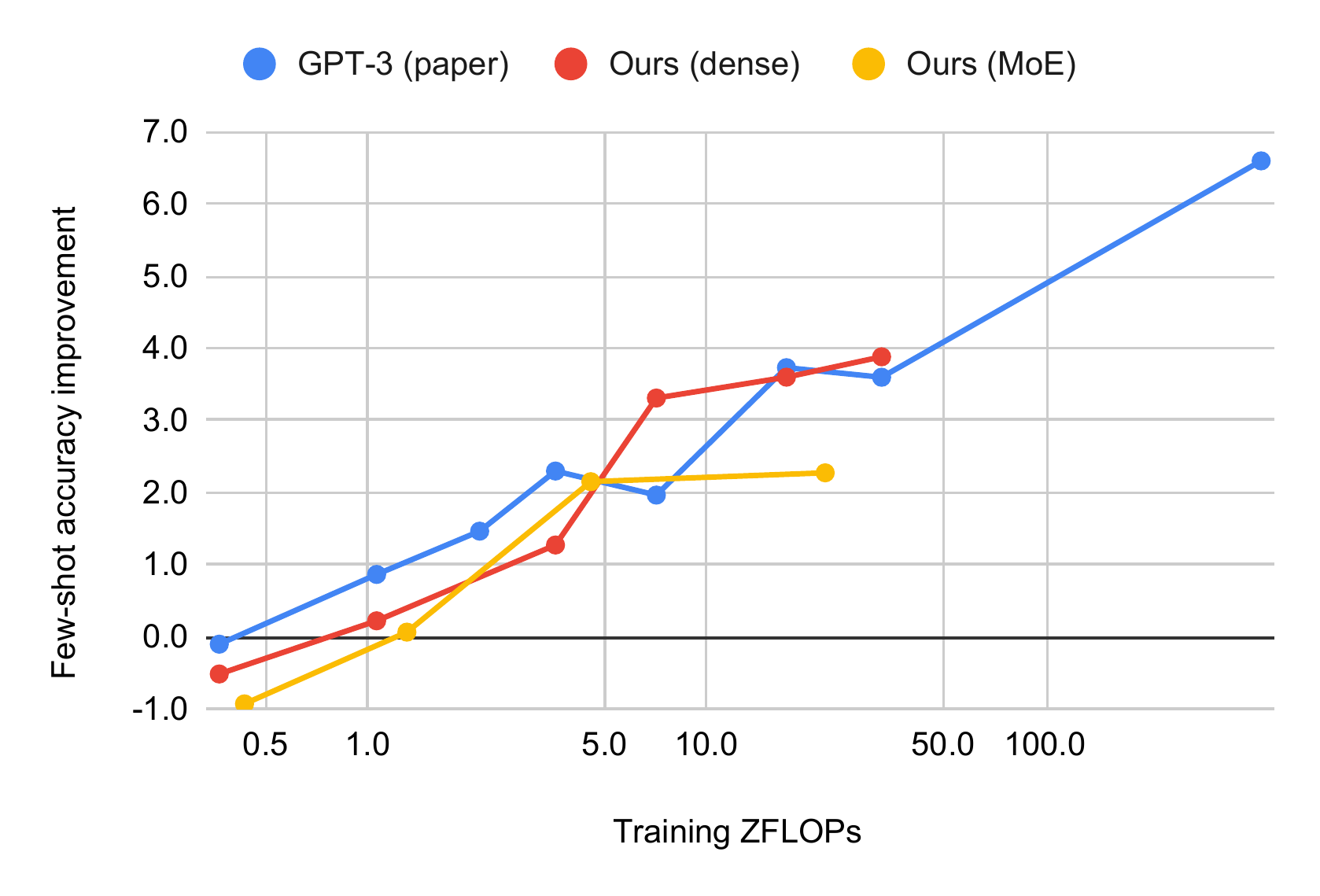}
\caption{
\textbf{Absolute accuracy improvement going from zero-shot to few-shot}, averaged across 3 tasks. Each point corresponds to a different, fully-trained model (see Table \ref{tab:models}).
\textit{GPT-3 (paper)} results taken from \citet{brown2020gpt3}.
}
\label{fig:fewshot}
\end{figure}

We report our few-shot results in Table \ref{tab:fewshot} and plot the corresponding improvement over zero-shot in Figure  \ref{fig:fewshot}.

Our dense baselines perform at par or slightly better than GPT-3. We observe that the improvement over zero-shot is bigger for larger models, further supporting that certain capabilities in language models emerge at scale \citep{brown2020gpt3}. Finally, we find that our larger MoE models also benefit from few-shot learning, outperforming their dense counterparts in all conditions. However, the improvements going from zero-shot to few-shot are smaller for MoE models compared to their dense counterparts. For example, the average for the 6.7B dense model improves by 3.6 points to 69.3 going from zero-shot to few-shot, whereas the corresponding 1.1T model improves by 2.3 points yielding 70.1.

\begin{table}[t!]
\begin{center}
\begin{small}
\addtolength{\tabcolsep}{-2pt}
\resizebox{0.48\textwidth}{!}{
\begin{tabular}{cr|ccc|c}
\toprule
&& WG & SC & OB & avg \\
\midrule
\multirow{8}{*}{\shortstack{GPT-3 \\ (paper)}}
& {125M} & 51.3 \scriptsize{--0.7} & 62.3 \scriptsize{--1.0} & 37.0 \scriptsize{+1.4} & 50.2 \scriptsize{--0.1} \\
& {355M} & 52.6 \scriptsize{+0.5} & 70.2 \scriptsize{+1.7} & 43.6 \scriptsize{+0.4} & 55.5 \scriptsize{+0.9} \\
& {760M} & 57.5 \scriptsize{+0.1} & 73.9 \scriptsize{+1.5} & 48.0 \scriptsize{+2.8} & 59.8 \scriptsize{+1.5} \\
& {1.3B} & 59.1 \scriptsize{+0.4} & 76.1 \scriptsize{+2.7} & 50.6 \scriptsize{+3.8} & 61.9 \scriptsize{+2.3} \\
& {2.7B} & 62.6 \scriptsize{+0.3} & 80.2 \scriptsize{+3.0} & 55.6 \scriptsize{+2.6} & 66.1 \scriptsize{+2.0} \\
& {6.7B} & 67.4 \scriptsize{+2.9} & 81.2 \scriptsize{+3.5} & 55.2 \scriptsize{+4.8} & 67.9 \scriptsize{+3.7} \\
& {13B} & 70.0 \scriptsize{+2.1} & 83.0 \scriptsize{+3.5} & 60.8 \scriptsize{+5.2} & 71.3 \scriptsize{+3.6} \\
& {175B} & 77.7 \scriptsize{+7.5} & 87.7 \scriptsize{+4.5} & 65.4 \scriptsize{+7.8} & 76.9 \scriptsize{+6.6} \\
\midrule
\multirow{6}{*}{\shortstack{Ours \\ (dense)}}
& {125M} & 52.2 \scriptsize{+0.1} & 64.7 \scriptsize{--1.3} & 35.0 \scriptsize{--0.4} & 50.7 \scriptsize{--0.5} \\
& {355M} & 53.7 \scriptsize{--0.5} & 72.2 \scriptsize{+1.1} & 42.0 \scriptsize{+0.0} & 56.0 \scriptsize{+0.2} \\
& {1.3B} & 60.1 \scriptsize{+2.0} & 78.6 \scriptsize{+1.9} & 49.4 \scriptsize{+0.0} & 62.7 \scriptsize{+1.3} \\
& {2.7B} & 63.9 \scriptsize{+2.5} & 82.1 \scriptsize{+3.8} & 53.2 \scriptsize{+3.6} & 66.4 \scriptsize{+3.3} \\
& {6.7B} & 67.6 \scriptsize{+2.9} & 83.2 \scriptsize{+2.7} & 57.0 \scriptsize{+5.2} & 69.3 \scriptsize{+3.6} \\
& {13B} & 71.0 \scriptsize{+3.5} & 85.0 \scriptsize{+4.1} & 59.5 \scriptsize{+4.1} & 71.8 \scriptsize{+3.9} \\
\midrule
\multirow{4}{*}{\shortstack{Ours \\ (MoE)}}
& {15B} & 52.5 \scriptsize{--0.9} & 71.4 \scriptsize{--2.1} & 42.2 \scriptsize{+0.2} & 55.4 \scriptsize{--0.9} \\
& {52B} & 58.1 \scriptsize{+0.7} & 77.5 \scriptsize{+1.6} & 48.9 \scriptsize{--2.1} & 61.5 \scriptsize{+0.1} \\
& {207B} & 62.8 \scriptsize{+1.9} & 81.1 \scriptsize{+3.0} & 52.4 \scriptsize{+1.6} & 65.4 \scriptsize{+2.2} \\
& {1.1T} & 68.6 \scriptsize{+2.3} & 83.9 \scriptsize{+2.1} & 57.7 \scriptsize{+2.5} & 70.1 \scriptsize{+2.3} \\
\bottomrule
\end{tabular}
}
\end{small}
\end{center}
\caption{
\textbf{Few-shot priming accuracy and absolute improvement over zero-shot.}
\textit{GPT-3 (paper)} results taken from \citet{brown2020gpt3}, all the other results were obtained by us as described in \S\ref{subsec:downstream}.
\texttt{WG}: WinoGrande, \texttt{SC}: StoryCloze, \texttt{OB}: OpenBookQA.}
\label{tab:fewshot}
\end{table}

\subsubsection{Supervised Fine-Tuning}
Table \ref{tab:finetuning} contrasts full fine-tuning performance of MoE models with their dense counterparts on 8 datasets, using zero-shot accuracy as a baseline for reference. We did not fine-tune the 6.7B and 13B dense models and the 1.1T MoE models, owing to their high resource needs. As expected, supervised fine-tuning yields substantial performance benefits for all dense models across all datasets, over zero-shot performance. In contrast, although fine-tuning of MoE models produces substantial benefits for Storycloze, BoolQ, SST-2, MNLI and some improvements on OpenBookQA, it results in worse performance for HellaSwag, PIQA, and Winogrande. For the cases where we see improvements, the accuracy of fine-tuned MoE models approach that of their corresponding dense models. For this comparison, we fine-tune MoE models exactly as we do the dense models. While MoE models may benefit from alternative fine-tuning approaches, for example, selective fine-tuning of the expert or non-expert parameters, we leave such exploration to future work. 

\begin{table}[t!]
\begin{center}
\begin{small}
\addtolength{\tabcolsep}{-2.5pt}
\resizebox{0.48\textwidth}{!}{
\begin{tabular}{cr|cccc|ccc}
\toprule
&& \multicolumn{4}{c|}{Ours (Dense)} & \multicolumn{3}{c}{Ours (MoE)} \\
&& 125M & 355M & 1.3B & 2.7B & 15B & 52B & 207B \\ %
\midrule
\multirow{2}{*}{\texttt{SC}}
& zero-shot & 66.0 & 71.0 & 76.8 & 78.2 & 73.6 & 75.9 & 78.1 \\ %
& fine-tune & 87.8  & 89.5 & 93.8 & 97.0 & 80.3 & 84.9 & 80.9 \\ %
\midrule
\multirow{2}{*}{\texttt{OB}}
& zero-shot & 35.4 & 42.0 & 49.4 & 49.6 & 42.0 & 51.0 & 50.8 \\ %
& fine-tune & 50.6 & 59.0 & 67.4 & 70.8 & 51.2 & 51.4 & 51.0 \\ %
\midrule
\multirow{2}{*}{\texttt{BQ}}
& zero-shot & 56.1 & 58.6 & 58.7 & 60.3 & 60.9 & 56.0 & 54.2 \\ %
& fine-tune & 73.2 & 75.2 & 79.6 & 84.6 & 71.6 & 75.3 & 77.5 \\ %
\midrule
\multirow{2}{*}{\texttt{MN}}
& zero-shot & 46.2 & 52.1 & 55.3 & 56.0 & 49.3 & 52.1 & 52.6 \\ %
& fine-tune & 80.9 & 84.3 & 84.1 & 88.9 & 77.7 & 81.2 & 78.7 \\ %
\midrule
\multirow{2}{*}{\texttt{SST-2}}
& zero-shot & 50.9 & 50.9 & 51.6 & 51.9 & 51.6 & 50.9 & 50.9 \\ %
& fine-tune & 92.9 & 92.9 & 94.8 & 93.4 & 89.3 & 90.1 & 90.3 \\ %
\midrule
\multirow{2}{*}{\texttt{HS}}
& zero-shot & 33.7 & 46.2 & 58.4 & 65.9 & 53.2 & 64.9 & 70.5 \\ %
& fine-tune & 50.7 & 64.8 & 74.1 & 90.0 & 37.3 & 45.4 & 42.2 \\ %
\midrule
\multirow{2}{*}{\texttt{PI}}
& zero-shot & 65.3 & 70.6 & 74.6 & 76.6 & 74.3 & 76.8 & 78.2 \\ %
& fine-tune & 68.2 & 71.7 & 71.2 & 80.3 & 66.3 & 66.1 & 68.3 \\ %
\midrule
\multirow{2}{*}{\texttt{WG}}
& zero-shot & 52.1 & 54.2 & 58.1 & 61.4 & 53.4 & 57.4 & 60.9 \\ %
& fine-tune & 65.7 & 63.3 & 67.4 & 69.5 & 50.2 & 56.0 & 50.4 \\ %
\bottomrule
\end{tabular}}
\end{small}
\end{center}
\caption{
\textbf{Fully supervised fine-tuning accuracy compared with zero-shot accuracy.}  \texttt{SC}: StoryCloze, \texttt{OB}: OpenBookQA, \texttt{BQ}: BoolQ, \texttt{MN}: MNLI, \texttt{HS}: HellaSwag, \texttt{PI}: PIQA, \texttt{WG}: WinoGrande. Largest models omitted owing to their high resource utilization.}
\label{tab:finetuning}
\end{table}

\section{Conclusion}

We present results for scaling sparse Language Models up to 1.1T parameters. We observe that up to this scale sparse models offer better performance vs. computation trade-off when compared to their dense counterparts for language modeling, zero- and few-shot learning. While the gap begins to close at scale our biggest sparse model outperforms its dense counterpart where the latter requires twice as much computation. These results confirm that sparse MoE models can provide an alternative to widely used dense architectures that saves computation and reduces model energy consumption.

\section*{Ethical considerations}
Previous work~\cite{sheng2019woman,bordia2019identifying,nadeem2020stereoset,de2021stereotype} has observed that language models absorb bias and toxicity represented in the training data. So as to better understand the potential harms of our models in this front, we evaluated them on StereoSet~\cite{nadeem2020stereoset} and CrowS-Pairs~\cite{nangia-etal-2020-crows}, and report our results in Appendix \ref{app:potential_harms}. Our results show that the percentage of bias and stereotype in dense and MoE models is comparable, especially at scale.
Moreover, in general, we note worse performance (more bias/stereotyping) at larger scales. This observation points to more research needed in order to mitigate such behavior. Intuitively however, we believe that sparse models may be inherently more controllable -- e.g.~designing specific experts -- than dense models. We leave this line of investigation for future research. 

Another concern of scaling language models is the energy usage and the associated environmental impact required for training, which we discuss in detail in Appendix \ref{app:co2}. Nevertheless, our work shows that MoEs can be a more compute-efficient alternative to traditional dense models, which could alleviate the environmental impact of future scaling efforts.
Moreover, by releasing all of our pre-trained language models, we believe we have alleviated some exploration burden for the community and the environment, allowing for more efficient offsets for other researchers.

In the spirit of transparency and allowing for maximal replicability and accountability, we include data and model cards together with our code.

\section*{Limitations}

Our study is limited to one specific MoE configuration. In particular, our sparse and dense models use the same hyperparameters and model structure, closely following GPT-3. However, it is possible that this configuration is suboptimal for scaling MoE models. Similarly, we did not explore different MoE-specific hyperparameters, such as the number of experts. Finally, while our work reveals that the performance gap between MoE and dense models varies greatly across tasks and domains, the specific factors that make certain tasks more favorable for MoEs remain unclear.

\bibliography{anthology,custom}
\bibliographystyle{acl_natbib}

\clearpage

\appendix

\begin{table*}[t!]
\begin{center}
\begin{small}
\addtolength{\tabcolsep}{-2.5pt}
\begin{tabular}{cllrrrrrrlrrrr}
\toprule
&&& \multicolumn{6}{c}{Dense} && \multicolumn{4}{c}{MoE} \\
\cmidrule{4-9}
\cmidrule{11-14}
&&& 125M & 355M & 1.3B & 2.7B & 6.7B & 13B && 15B & 52B & 207B & 1.1T \\
\midrule
In-domain & Validation && 20.65 & 15.14 & 12.48 & 10.92 & 9.82 & 8.97 &  & 12.58 & 9.58 & 8.76 & 6.80 \\
\midrule
\multirow{23}{*}{\shortstack{Out-of-domain \\ --- \\ (The Pile)}}
& ArXiv &  & 15.74 & 11.42 & 9.00 & 8.03 & 7.29 & 6.86 &  & 10.81 & 8.79 & 7.72 & 6.91 \\
& Bibliotik &  & 26.78 & 19.62 & 15.75 & 13.96 & 12.81 & 11.96 &  & 17.80 & 14.64 & 13.29 & 11.88 \\
& BookCorpus &  & 23.60 & 17.91 & 14.83 & 13.36 & 12.38 & 11.70 &  & 16.54 & 13.90 & 12.82 & 11.57 \\
& CommonCrawl &  & 22.49 & 16.92 & 13.91 & 12.47 & 11.50 & 10.80 &  & 15.17 & 12.47 & 11.43 & 10.02 \\
& DM\_Mathematics &  & 12.29 & 9.73 & 8.51 & 8.10 & 7.66 & 7.41 &  & 10.51 & 8.82 & 8.28 & 7.63 \\
& Enron\_Emails &  & 19.98 & 15.71 & 12.32 & 11.40 & 10.78 & 10.09 &  & 14.24 & 12.18 & 11.08 & 10.45 \\
& EuroParl &  & 27.16 & 15.80 & 12.02 & 9.91 & 8.63 & 7.68 &  & 12.58 & 9.47 & 8.41 & 6.92 \\
& FreeLaw &  & 16.78 & 11.98 & 9.44 & 8.33 & 7.58 & 7.08 &  & 10.54 & 8.49 & 7.68 & 6.84 \\
& Github &  & 8.92 & 6.55 & 5.13 & 4.61 & 4.30 & 4.03 &  & 6.11 & 4.93 & 4.41 & 3.99 \\
& Gutenberg\_PG-19 &  & 29.15 & 20.70 & 16.39 & 14.37 & 13.03 & 12.08 &  & 18.85 & 14.94 & 13.48 & 11.90 \\
& HackerNews &  & 29.37 & 22.53 & 18.11 & 16.20 & 14.96 & 14.08 &  & 20.72 & 17.15 & 15.67 & 14.21 \\
& NIH\_ExPorter &  & 26.78 & 19.18 & 15.28 & 13.55 & 12.40 & 11.64 &  & 17.08 & 14.00 & 12.66 & 11.42 \\
& OpenSubtitles &  & 20.72 & 16.87 & 14.16 & 13.05 & 12.28 & 11.61 &  & 16.38 & 13.64 & 12.64 & 11.78 \\
& OpenWebText2 &  & 20.56 & 14.64 & 11.88 & 10.47 & 9.51 & 8.79 &  & 12.76 & 10.03 & 9.03 & 7.06 \\
& PhilPapers &  & 27.60 & 20.00 & 15.99 & 14.07 & 12.90 & 12.03 &  & 18.77 & 15.19 & 13.56 & 11.91 \\
& PubMed\_Abstracts &  & 24.16 & 16.56 & 12.95 & 11.38 & 10.35 & 9.65 &  & 14.46 & 11.71 & 10.49 & 9.36 \\
& PubMed\_Central &  & 12.19 & 9.35 & 7.65 & 6.89 & 6.36 & 6.03 &  & 8.74 & 7.32 & 6.60 & 6.02 \\
& StackExchange &  & 17.76 & 12.46 & 9.65 & 8.43 & 7.58 & 7.04 &  & 10.99 & 8.74 & 7.73 & 6.72 \\
& USPTO &  & 17.15 & 12.85 & 10.43 & 9.36 & 8.67 & 8.18 &  & 12.00 & 9.95 & 9.01 & 8.14 \\
& Ubuntu\_IRC &  & 28.40 & 21.47 & 16.16 & 13.92 & 12.50 & 11.48 &  & 17.80 & 14.79 & 12.85 & 11.49 \\
& Wikipedia\_en &  & 20.51 & 14.61 & 11.59 & 10.22 & 9.22 & 8.56 &  & 12.35 & 9.68 & 8.60 & 6.65 \\
& YoutubeSubtitles &  & 19.01 & 13.51 & 10.80 & 9.31 & 8.38 & 7.70 &  & 12.23 & 9.88 & 8.80 & 7.51 \\
\cmidrule{2-14}
& Average && 21.23 & 15.47 & 12.36 & 10.97 & 10.05 & 9.39 && 13.97 & 11.40 & 10.28 & 9.11 \\
\bottomrule
\end{tabular}
\end{small}
\end{center}
\caption{
\textbf{Full perplexity results}.}
\label{tab:ppl}
\end{table*}

\section{Full perplexity results}\label{app:full-results}

Table \ref{tab:ppl} reports the full perplexity results, including all the different subsets of the Pile.

\section{Fine-tuning Settings}
\label{sec:fine_tuning_settings}
We run fine-tuning for a fixed number of epochs (100 for BoolQ, OpenBookQA, StoryCloze, and PIQA, 25 for HellaSwag, Winogrande, and SST-2, 6 for MNLI) and perform model selection based on validation set accuracy. For datasets either without a validation set or where we evaluate on the validation set, we randomly split the training set and use 80\% for fine-tuning and 20\% for per-epoch validation.

\section{Understanding Potential Harms}\label{app:potential_harms}
Previous work~\cite{sheng2019woman,bordia2019identifying,nadeem2020stereoset,de2021stereotype} has observed that language models absorb bias and toxicity represented in the training data. %
We set out to explore if sparse models would behave differently than dense models in this arena.
To that end, we evaluate our dense and MoE models on two popular benchmarks: StereoSet~\cite{nadeem2020stereoset} and CrowS-Pairs~\cite{nangia-etal-2020-crows}. StereoSet measures bias across four domains: \emph{profession}, \emph{gender}, \emph{religion}, and \emph{race}. CrowS-Pairs dataset covers nine bias types: \emph{race}, \emph{gender/gender identity}, \emph{sexual orientation}, \emph{religion}, \emph{age}, \emph{nationality}, \emph{disability}, \emph{physical appearance}, and \emph{socioeconomic status}.\footnote{The two benchmarks have limitations such as lack of clear articulations of how certain biases are being measured \cite{blodgett2021stereotyping}. Results should be interpreted accordingly.}%

\paragraph{Stereotypical bias as a function of scale.}
Table~\ref{tab:stereoset-results} presents the results on StereoSet benchmark using three metrics: (1) \emph{Language Modeling Score (LMS)}: defined as the percentage of instances in which a language model prefers meaningful over meaningless associations (higher LMS is better); (2) \emph{Stereotype Score (SS)}: defined as the percentage of instances where a model prefers a stereotypical association over an anti-stereotypical association (SS score close to $50$ is better, while a more biased model would have a higher score towards $100\%$); (3) Idealized CAT Score (ICAT): defined as a combination of LMS and SS to capture both in a single metric: ${\scriptstyle LMS} * \frac{min(SS,  100-SS)}{50}$ (higher ICAT is better).

Table~\ref{tab:crowspairs-results} presents the performance of our models on CrowS-Pairs using the \emph{Stereotype Score (SS)} metric. %
Similar to StereoSet, we observe that both dense and MoE models get worse with scale, again with statistically significant ($p<0.05$) difference between best and worst scores based on bootstrap test~\cite{noreen1989computer,efron1994introduction}.

\begin{table}[t!]
\begin{center}
\begin{small}
\addtolength{\tabcolsep}{-4pt}
\resizebox{0.95\linewidth}{!}{
\begin{tabular}{cr|ccccc|cccc}
\toprule
&& \multicolumn{5}{c|}{Ours (Dense)} & \multicolumn{4}{c}{Ours (MoE)} \\
Category &
& 125M & 355M & 1.3B & 2.7B & 6.7B & 15B & 52B & 207B & 1.1T \\ 
\midrule
\multirow{3}{*}{Prof.}
& \texttt{LMS} & 80.8 & 82.0 & 81.9 & 80.8 & 79.3 & 79.7 & 80.5 & 81.1& 78.0 \\ 
& \texttt{SS} & 48.2 & 49.8 & 51.3 & 53.6 & 54.2 & 52.2 & 54.9 & 54.8 & 54.4\\ 
& \texttt{ICAT} & 77.9 & 81.7 & 79.8 & 75.1 & 72.6 & 76.2 & 72.6 & 73.4 & 71.1 \\ 
\midrule
\multirow{3}{*}{Gender}
& \texttt{LMS} & 83.3 & 82.4 & 83.9 & 83.1 & 82.2 & 81.4 & 82.9 & 82.2 & 80.2 \\ 
& \texttt{SS} & 59.9 & 59.1 & 59.1 & 60.7 & 60.3 & 58.7 & 58.3 & 58.7 & 61.2 \\ 
& \texttt{ICAT} & 66.7 & 67.4 & 68.6 & 65.2 & 65.2 & 67.3 & 69.2 & 68.0 & 62.3 \\
\midrule
\multirow{3}{*}{Reli.}
& \texttt{LMS} & 85.9 & 87.8 & 87.2 & 87.2 & 85.3 & 87.8 & 85.9 & 83.3 & 81.4\\ 
& \texttt{SS} &  50.0 & 46.2 & 50.0 & 55.1 & 52.6 & 50.0 & 48.7 & 51.3 & 51.3 \\ 
& \texttt{ICAT} & 85.9 & 81.1 & 87.2 & 78.2 & 80.9 & 87.8 & 83.7 & 81.2 & 79.3 \\
\midrule
\multirow{3}{*}{Race}
& \texttt{LMS} & 82.3 & 82.3 & 83.8 & 83.0 & 83.1 & 83.7 & 83.4 & 82.0 & 82.1\\ 
& \texttt{SS} &  42.9 & 45.7 & 48.3 & 49.8 & 50.3 & 47.1 & 47.3 & 49.7 & 47.5\\ 
& \texttt{ICAT} &  70.7 & 75.2 & 80.8 & 82.7 & 82.6 & 78.9 & 79.0 & 81.5 & 78.0\\
\midrule
\multirow{3}{*}{Overall}
& \texttt{LMS} & 82.0 & 82.4 & 83.2 & 82.3 & 81.6 & 82.1 & 82.3 & 81.7 & 80.2\\ 
& \texttt{SS} &  47.2 & 48.8 & 50.7 & 52.7 & 53.0 & 50.5 & 51.6 & 52.8 & 51.9 \\ 
& \texttt{ICAT} & 77.4 & 80.5 & 82.0 & 77.9 & 76.6 & 81.2 & 79.7 & 77.2 & 77.2 \\

\bottomrule
\end{tabular}
}
\end{small}
\end{center}
\caption{\textbf{Stereotypical bias comparison on the inter-sentence task of StereoSet.} Note that for LMS and ICAT higher is better whereas for SS closer to $50$ is better. All scores are macro averages of all samples present in a category. Overall represents all the samples in this dataset.}
\label{tab:stereoset-results}
\end{table}

\begin{table}[t!]
\begin{center}
\begin{small}
\addtolength{\tabcolsep}{-4pt}
\resizebox{\linewidth}{!}{
\begin{tabular}{l|ccccc|cccc}
\toprule
& \multicolumn{5}{c|}{Ours (Dense)} & \multicolumn{4}{c}{Ours (MoE)} \\
Category & 125M & 355M & 1.3B & 2.7B & 6.7B & 15B & 52B & 207B & 1.1T \\
\midrule
Gender & 56.5 & 58.4 & 56.5 & 58.0 & 60.3 & 56.9 & 63.4 & 62.2 & 60.7\\
Religion & 63.8 & 67.6 & 68.6 & 70.5 & 73.3 & 64.8 & 68.6 & 70.5 & 73.3 \\
Race/Color & 61.1 & 58.7 & 63.8 & 62.0 & 67.6 & 57.4 & 59.7 & 60.5 & 61.8 \\
Sexual orientation & 78.6 & 78.6 & 82.1 & 76.2 & 79.8 & 73.8 & 78.6 & 78.6 & 76.2 \\
Age & 57.5 & 60.9 & 60.9 & 62.1 & 62.1 & 66.7 & 59.8 & 66.7 & 66.7 \\
Nationality & 46.5 & 47.2 & 61.0 & 54.7 & 59.1 & 52.2 & 56.6 & 61.6 & 57.2 \\
Disability & 66.7 & 70.0 & 75.0 & 71.7 & 70.0 & 75.0 & 73.3 & 75.0 & 76.7 \\
Physical appearance & 73.0 & 65.1 & 69.8 & 71.4 & 74.6 & 71.4 & 71.4 & 74.6 & 79.4\\
Socioeconomic status & 69.2 & 66.9 & 72.7 & 68.0 & 71.5 & 69.8 & 69.8 & 72.1 & 73.3\\
\midrule
Overall & 61.3 & 60.9 & 65.1 & 63.4 & 67.0 & 61.4 & 63.9 & 65.5 & 65.7 \\
\bottomrule
\end{tabular}
}
\end{small}
\end{center}
\caption{\textbf{Stereotypical bias comparison on CrowS-Pairs.} Scores closer to $50$ are better. All scores are macro averages. }
\label{tab:crowspairs-results}
\end{table}

\paragraph{Stereotypical bias in dense vs MoE models.}
We observe that as the model size increases, both dense and MoE models get worse ICAT scores in general -- they become more biased with a statistically significant difference between best and worst scores. On the StereoSet benchmark, corresponding dense and sparse models (comparable FLOPs) yield comparable performance. On the CrowS-Pairs MoE models perform slightly better (less biased) than dense models on average but the difference is not statistically significant (see Table~\ref{tab:stereoset-results} and Table~\ref{tab:crowspairs-results}).

\section{CO2 Emission Related to Experiments}\label{app:co2}

\begin{table}[t]
\begin{center}
\begin{tabular}{lcc}
\toprule
& GPU days & tCO$_2$e \\
\midrule
125M dense & 26 & 0.1 \\
355M dense & 77 & 0.2 \\
1.3B dense & 258 & 0.8 \\
2.7B dense & 512 & 1.7 \\
6.7B dense & 1238 & 4.0 \\
13B dense & 2363 & 7.7 \\
\midrule
15B MoE & 43 & 0.1 \\
52B MoE & 131 & 0.4 \\
207B MoE & 456 & 1.5 \\
1.1T MoE & 2241 & 7.3 \\
\midrule
\textbf{Total} & 7345 & 23.8 \\
\midrule
\midrule
\multicolumn{2}{l}{GShard 600B MoE} & 4.8 \\
\multicolumn{2}{l}{Switch Transformer 1.5T MoE} & 72.2 \\
\multicolumn{2}{l}{GPT-3 175B} & 552.1 \\
\bottomrule
\end{tabular}
\end{center}
\caption{
\textbf{Estimated training time and energy costs to train the models reported in this paper}, based on the number of A100 GPU-days required for training.
See Appendix \ref{app:co2} for more details about these estimates.
}
\label{tab:energy}
\end{table}

The carbon emissions of the experiments reported in this work are dominated by the largest models, in particular the 13B parameter dense and 1.1T parameter MoE models.
We trained our largest models on Azure NDv4 instances\footnote{Some of our smaller models were trained on an on-premises cluster, but our estimates assume that all training was done on Azure for simplicity.} with A100 GPUs (TDP of 400W) in the West US 2 region, which has a carbon efficiency of 0.3 kgCO$_2$e/kWh and assumed Power Usage Effectiveness (PUE) of 1.125.\footnote{Carbon efficiency is estimated using the Machine Learning Impact calculator~\citep{lacoste2019quantifying} and the PUE estimate is based on \citet{bergman2021azure}.}
Thus each GPU-day of training is responsible for 3.24 kgCO$_2$e of emissions, of which 100 percent is directly offset by the cloud provider.

In Table~\ref{tab:energy} we report training time and energy usage for our models based on the above estimates, as well as estimates from \citet{patterson2021carbon} for other large-scale LMs, in particular GShard~\citep{lepikhin2021gshard}, Switch Transformer~\citep{fedus2021switch} and GPT-3~\citep{brown2020gpt3}.
Training times (GPU days) are computed assuming a throughput of 160 TFLOP/s and 115 TFLOP/s per A100 GPU for our dense and MoE models, respectively, based on observed training speeds for our largest models.\footnote{MoE models have additional all-to-all communication overhead, causing them to achieve lower GPU utilization compared to dense models. This overhead could be reduced with further optimization of the implementation.}

We note that these estimates do not account for the costs associated with manufacturing the infrastructure to train such models, which can be significant~\citep{gupta2021chasing,wu2021sustainable}.
We also note that these estimates do not account for pilot experiments common in the early exploratory stages of a research project.
We estimate that pilot experimentation adds a factor of 2 to the total training cost, since most exploration and tuning is performed at small scale where compute costs are small, and the largest models are typically trained at most once or twice.
For instance, we trained and discarded a pilot 6.7B dense and 1.1T MoE model in the early stages of this project, but trained the 13B dense model once.

\section{Knowledge Distillation}

\begin{table}[t]
\centering
\begin{tabular}{lcc}
\toprule
Teacher size & Cost & PPL \\
\midrule
None (Baseline) & 0.36 & 16.01 \\
MoE 15B         & 0.43 & 15.20 \\
355M Dense      & 1.06 & 14.88 \\
1.3B Dense      & 3.57 & 14.67 \\
MoE 52B         & 1.30 & 14.64 \\
2.7B Dense      & 7.08 & 14.62 \\
\bottomrule
\end{tabular}
\caption{\textbf{Distillation Results.} \emph{Cost} refers to the cost to train the teacher model (see Table~\ref{tab:models}). \emph{PPL} is the in-domain validation perplexity.}
\label{tab:distill}
\end{table}

In Section~\ref{subsubsec:efficiency-gain} we show that sparse (MoE) models are significantly more efficient to train than dense models.
However, inference for large sparse models can be challenging, since the large number of parameters (most of which are inactive) introduce significant storage costs compared to dense models.

In this section we explore whether it is possible to blend the benefits of dense and sparse models via knowledge distillation~\citep{hinton2015distilling}.
Building on recent work in this area~\citep{shleifer2020pretrained, sanh2020distilbert, fedus2021switch}, we train small dense ``student'' models to mimic the behavior of larger ``teacher'' models, which may be either large dense or large sparse (MoE) models.

\paragraph{Methods}
We train dense student models with 12 layers and hidden dimension 768, matching the 125M dense model architecture in Table~\ref{tab:models}.
We use a weighted training objective that combines the standard cross entropy loss (25\% weight) with a soft distillation loss (75\% weight) that encourages the student model to reproduce the logits of the teacher.
Additionally, we use a reduced sequence length of 1024 tokens to speed up experimentation.

\paragraph{Results}

We report results in Table~\ref{tab:distill}.
We find that student models trained with knowledge distillation improve over a well tuned dense baseline for both dense and sparse teacher models.
Furthermore, some of the efficiency advantages of sparse training can be transmitted to a dense student through distillation.
For example, student models distilled from a 52B parameter MoE teacher outperform student models distilled from a 1.3B parameter dense teacher, despite that the dense teacher model is twice as costly to train.

\section{Techniques for Large-scale Training}
\label{app:scaling}

We adopt several techniques to train models in this work, including a more memory-efficient recipe for FP16 training, activation checkpointing and Fully Sharded Data Parallel.

\paragraph{FP16 Training:}
Typical mixed-precision training recipes require storing model weights in both 16-bit (FP16) and 32-bit (FP32), as well as storing optimizer state in 32-bit to preserve accurate weight updates~\citep{micikevicius2018mixed,ott2018scaling}. Thus training a model in mixed precision with Adam requires 16 bytes of memory per parameter to maintain 16-bit and 32-bit weights (6 bytes), Adam optimizer state (8 bytes) and gradients (2 bytes), not including any memory required for activations.

In practice we find we can reduce memory requirements by 50\% by maintaining only 16-bit model weights, optimizer state and gradients with no loss in model accuracy.
First, we simply discard the 32-bit model weights, saving 4 bytes per parameter, since pilot experiments showed this to have no impact on model quality when training with large batch sizes.
Second, the Adam optimizer state can be stored in 16-bit by dynamically rescaling the values to avoid underflow.
Specifically, we compute the standard Adam weight update in 32-bit and then apply the following transformation at the end of each optimization step to maintain the optimizer state in 16-bit~\citep{dhariwal2020jukebox}:
\[
\vec{m}_{fp16} = \frac{\vec{m}}{\frac{\max abs(\vec{m})}{\textrm{FLOAT16\_MAX}} + \epsilon}
\]
where $\epsilon=10^{-8}$ and $\textrm{\small{FLOAT16\_MAX}}=65504.0$ is the largest finite value expressible in FP16.
We apply this transformation separately for the first and second moment estimates in Adam.

\paragraph{Activation Checkpointing:}
Activation size grows proportionally to the model and batch size, making it infeasible to store activations for transformer models with more than a couple of billion parameters.
We adopt a popular technique called \emph{activation checkpointing}, which saves memory during training by discarding a subset of activations in the forward pass and recomputing them in the backward pass~\citep{chen2016training}.
This technique results in a 33\% increase in computation, but can often reduce activation memory requirements by a factor of 10~\citep{rajbhandari2020zero}.
In our experiments we only store activations between transformer layers and recompute intermediate activations within each layer during the backward pass.

\paragraph{Fully Sharded Data Parallel:}
In data parallel training, gradients are averaged across multiple workers (GPUs) that process distinct partitions of the data.
Standard implementations maintain redundant copies of the model weights and optimizer state on each GPU, however this wastes GPU memory and makes it challenging to scale the model size beyond what can fit on a single GPU.
Recent work has explored sharding model parameters, optimizer state and gradients across workers~\citep{xu2020automatic,rajbhandari2020zero}, enabling training of models with more than one trillion parameters using only data parallelism without the added complexity introduced by model parallel training approaches like pipeline or tensor parallelism~\citep{huang2019gpipe,shoeybi2019megatron,narayanan2021efficient}.

We implement these ideas in \emph{Fully Sharded Data Parallel (FSDP)},\footnote{Fully Sharded Data Parallel is a drop-in replacement for PyTorch's Distributed Data Parallel module and is available at \url{github.com/facebookresearch/fairscale}.}
which shards model parameters in-place and gathers the parameters on all workers just-in-time for the forward and backward pass.
Training with FSDP is typically faster than standard data parallel implementations for three reasons: (1) sharding reduces the cost of the optimizer step and weight update by distributing it across workers, rather than redundantly updating model replicas on each worker; (2) while FSDP introduces 50\% more communication, this extra communication is overlapped with the computation in the forward and backward pass; and (3) FSDP yields significant memory savings, which can be used to increase the batch size and achieve higher GPU utilization.

One important decision when using FSDP is choosing which submodules in the model to ``wrap" with FSDP.
If the wrapping is too fine-grained, then the parameter shards will be very small which reduces communication efficiency.
If the wrapping is too coarse, then this increases the peak resident memory and may pose challenges when scaling to larger model sizes.
In this work we wrap every transformer layer with FSDP, which ensures a reasonably large message size for communication while still limiting the peak resident memory to the size of a single layer.

\section{Counting FLOPs}
\label{app:flops}

We count the number of floating-point operations (FLOPs) analytically following \citet{narayanan2021efficient}.
We assume that all models are trained with activation checkpointing and thus have an additional forward pass before the backward pass.
Thus the total training FLOPs for our dense models is given by:
\[
F_{\textrm{dense}} = 96Tlh^2\left(1 + \frac{s}{6h} + \frac{V}{16lh}\right),
\]
where $T$ is the total number of training tokens, $l$ is the number of layers, $h$ is the hidden dimension, $s$ is the sequence length and $V$ is the vocabulary size. In this work, $T=300e^9$, $s=2048$ and $V=51200$ for all models.

For mixture of expert models, we account for an additional feed-forward network at every other layer for the top-2 routing in GShard~\cite{lepikhin2021gshard}, and ignore the FLOPs of the routing projection which is negligible.
The resulting training FLOPs for our MoE models is given by:
\[
F_{\textrm{MoE}} = F_{\textrm{dense}} + 32Tlh^2.
\]
Notably, this quantity is independent of the number of experts.

\end{document}